\documentclass{article}

\usepackage{microtype}
\usepackage{graphicx}
\usepackage{subfigure}          
\usepackage{booktabs} % for professional tables
\usepackage{float}

\usepackage{algorithmic}

\usepackage[accepted]{icml2022}

\icmltitlerunning{Linear Time Kernel Matrix Approximation via Hyperspherical Harmonics}

\usepackage{amssymb}
\usepackage{amsmath}
\usepackage{mathtools}
\usepackage{empheq}
\usepackage{dsfont}
\usepackage{amsthm}
\usepackage{bbm}
\usepackage{enumerate}
\usepackage{bm}

\newcommand{\cU}{\mathcal{U}}

\newtheorem{remark}{Remark}
\DeclareMathOperator*{\argmin}{arg\,min}

\begin{document}

\twocolumn[
% \icmltitle{AutoFMM}
% \icmltitle{AutoFMM: A Fast Method for Gaussian Processes}
% \icmltitle{AutoFMM: A Fast Algorithm for Kernel Methods}
% \icmltitle{AutoFMM: A Fast Algorithm for Kernel Machines}
% \icmltitle{AutoFMM: A Fast Algorithm for Kernel Matrix}
% \icmltitle{AutoFMM: A Fast Multiplication Algorithm for Kernel Matrices}
\icmltitle{Linear Time Kernel Matrix Approximation via Hyperspherical Harmonics}




\begin{icmlauthorlist}
\icmlauthor{John Paul Ryan}{cornell}
\icmlauthor{Anil Damle}{cornell}
\end{icmlauthorlist}

\icmlaffiliation{cornell}{Department of Computer Science, Cornell University, Ithaca, USA}

\icmlcorrespondingauthor{John Paul Ryan}{jpr269@cornell.edu}
\icmlkeywords{Machine Learning, ICML}

\vskip 0.3in
]



\printAffiliationsAndNotice{}  % leave blank if no need to mention equal contribution

\begin{abstract}
We propose a new technique for constructing low-rank approximations of matrices that arise in kernel methods for machine learning. 
Our approach pairs a novel automatically constructed analytic expansion of the underlying kernel function with a data-dependent compression step to further optimize the approximation. 
This procedure works in linear time and is applicable to any isotropic kernel. Moreover, our method accepts the desired error tolerance as input, in contrast to prevalent methods which accept the rank as input. 
Experimental results show our approach compares favorably to the commonly used Nystr\"{o}m method with respect to both accuracy for a given rank and computational time for a given accuracy across a variety of kernels, dimensions, and datasets. 
Notably, in many of these problem settings our approach produces near-optimal low-rank approximations.
We provide an efficient open-source implementation of our new technique to complement our theoretical developments and experimental results.
\end{abstract}

% and it is particularly efficient when the data dimension is moderate relative to the dataset size. 

% This procedure yields near-optimal low-rank approximations in linear time, is applicable to any isotropic kernel, and is particularly efficient when the data dimension is moderate relative to the dataset size. 

% Our approach is based on a novel data-dependent analytic expansion of the underlying kernel function which allows for linear time formation of a low-rank approximation with respect the the size of the dataset. Our method is applicable whenever the kernel is isotropic and the number of dimensions is moderate compared to the size of the dataset. 

\section{Introduction}
Kernel methods are a popular class of techniques in machine learning problems. They typically involve the implicit definition of an infinite dimensional feature space using an explicitly defined kernel function that describes similarity between data points. Examples of techniques based around the use of kernels include kernel support vector machines, kernel ridge regression, spectral clustering, kernel PCA, and Gaussian process regression~\citep{shawe2004kernel, scholkopf2018learning}. In practice, it is important that an appropriate kernel be chosen for the given task and a wealth of literature exists on proper choices of kernel functions and their hyperparameters.

In this work, our focus is on computational problems that generically arise in the use of kernel methods. Let $k$ be an isotropic kernel function and let $\mathcal{X}$ be a set of points in $\mathbb{R}^d$. The associated kernel matrix is defined as
\[
K_{ij} =k(\|x_i-x_j\|)\qquad x_i,x_j\in\mathcal{X}.
\]
Simply populating and storing the kernel matrix requires time and space which grow quadratically with the size of the dataset. Typically, kernel methods will also require operations on the kernel matrix which na\"{i}vely take cubic time, including solving linear systems and/or computing eigendecompositions. For even modestly sized datasets standard approaches are infeasible. This paper introduces a novel technique for constructing an approximation to the kernel matrix in linear time that subsequently enables the computation of matrix vector products with $K$ in linear time. This procedure can be used to, for example, substantially accelerate the solution of linear systems via iterative methods~\citep{iterbook, cgbrief}.

Accelerating computations involving kernel matrices has been the subject of extensive research. One family of algorithms uses inducing points as a means by which to approximately represent the kernel matrix in an easier-to-use form. One example is the Nystr\"{o}m method~\citep{nystrom1930praktische} which uses kernel interactions between data points and a set of inducing points as basis functions for a low-rank approximation of the matrix. For example, if $\mathcal{U}\subset \mathcal{X}$ is a subset of datapoints, then the Nystr\"{o}m approximation is given by
\begin{equation}
    \label{eqn:nystrom}
    K\approx K_{\text{Nys}} = K_{\mathcal{X}\cU}(K_{\cU\cU})^{-1}K_{\cU\mathcal{X}}
\end{equation}
This low rank approximation allows for accelerated matrix-vector multiplies within iterative schemes as well as the opportunity to use the Sherman-Morrison-Woodbury formula in settings where the kernel matrix has a diagonal matrix added (such as for regularization).

Other techniques leverage analytic properties of the kernel function itself to arrive at more manageable forms of the kernel matrix. For example, the Fast Multipole Method and Fast Gauss Transform~\citep{greengard1987fast,greengard1991fast,yang2003improved} both use truncated expansions of the kernel function to compute low-rank matrix representations without ever having to examine the original kernel matrix. The method of random Fourier features~\citep{rahimi2007random} uses a representation of the kernel as the expectation of a random function to approximate the kernel matrix via a Monte-Carlo scheme. Recent work~\citep{solin} leverages the spectral density of the kernel to compute coefficients in a harmonic expansion. 
Whereas the techniques of the previous paragraph use kernel interactions among data points without explicitly analyzing the kernel function itself, these analytically motivated methods use only properties of the kernel as the basis of their approximation. While data-independent approximations benefit from being easy to update with new datapoints, they often yield higher-rank representations than necessary. In addition, they are often kernel-targeted, requiring knowledge of a known expansion for a given kernel, or the kernel's spectral density.

\subsection{Contributions}
We present a new technique which leverages both the analytic properties of the underlying kernel (without being targeted to any specific kernel), and additional data-dependent compression to generate a low-rank approximation to the kernel matrix. The time and space cost of the approximation are linear in the size of the dataset. The main baseline technique with which we compare our results is the Nystr\"{o}m method for high-accuracy kernel-independent approximations. Our experiments show favorable comparison with the Nystr\"{o}m method for a host of kernels, dimensions, datasets, and error tolerances. In fact, we show a number of settings in which our method achieves a nearly-optimal tradeoff between rank and error, while still taking less time than the Nystr\"{o}m method. 

\section{Prior Work}
\label{sec:prior}

Given a kernel function $k$ and dataset $\mathcal{X}$ there are several common techniques for devising a fast matrix vector product with the associated kernel matrix $K.$ One broad class of methods computes low-rank approximations to $K$ as $K\approx UV^T$ when applicable. The SVD is the optimal choice if accuracy for a given rank is the only consideration. For sufficiently small scale problems this may be feasible. However, for larger problems computing the SVD becomes prohibitively expensive since, at a minimum, methods targeting the top singular values/vectors require repeated matrix vector products with $K$---the very problem we want to address. 

The Nystr\"{o}m method \cite{nystrom1930praktische} is a general approach to this problem that is much more computationally efficient at the expense of accuracy for a fixed rank approximation (i.e., it is suboptimal relative to the SVD). The most commonly used variants within machine learning are sampling based methods that yield low-rank approximations of $K$ as in~\eqref{eqn:nystrom}.
\footnote{The Nystr\"{o}m method was originally introduced within the integral equations community~\cite{nystrom1930praktische} and subsequently found use within machine learning~\cite{williams2001using,gittens2013revisiting}.} 
The key complexities of this methods are: (1) selecting an appropriate rank for the approximation and (2) selecting the subset $\cU.$ In principle, the theory of pseudoskeleton approximations~\cite{goreinov1997theory} provides generic upper bounds for the rank given a desired accuracy and there is significant work within the machine learning and theoretical computer science communities to sharply characterize the best way to select $\cU$ (see, e.g., \cite{musco2017recursive,gittens2013revisiting,drineas2005nystrom}). Another broad class of methods that fall into this category are so-called random features methods such as random Fourier features~\cite{rahimi2007random}. In this setting, the low-rank factors are produced by approximating the feature expansion related to a given kernel. While these methods are quite efficient, they often struggle to achieve high accuracy without taking the approximation to be of excessively large rank. For a systematic comparison of these methods see~\cite{yang2012nystrom}.

Another set of methods broadens these techniques to cases where the kernel matrix $K$ may not be globally low rank, but it exhibits so-called rank-structure. This is often captured using domain decomposition schemes which can leverage analytic expansions of kernel functions as in \cite{greengard1987fast, ryan2021fast}, or algebraic data-dependent methods as in \citep{ying2004kernel,ying2006kernel, meka}. These methods typically require the ability to compute low-rank approximations of subblocks of $K$\textemdash our present work solves this problem making it complementary to these schemes rather than in direct competition. 

Another class of methods for solving this problem represent $K$ as a product $K\approx WK_{\mathcal{G}\mathcal{G}}W^T,$ where $\mathcal{G}$ is a set of points chosen such that $K_{\mathcal{G}\mathcal{G}}$ admits fast matrix vector products and $W$ is an appropriately chosen interpolation/aggregation matrix. The most common set of methods in this class let $\mathcal{G}$ be a regular grid over the domain of interest and forcing $W$ to be sparse (i.e., using local interpolation). In this setting $K_{\mathcal{G}\mathcal{G}}$ can be efficiently applied using the fast Fourier transform. This form of method arose as the so-called pre-corrected FFT~\citep{phillips1994precorrected,white1994comparing} and gained popularity in the machine learning community as structured kernel interpolation (SKI)~\citep{wilson2015kernel} (in this context, it can be seen as an acceleration of inducing point methods~\citep{snelson2005sparse}). While potentially highly efficient in moderate dimension if sparse interpolation and aggregation operators are used (i.e., $W$ is very sparse), as with sampling based Nystr\"{o}m methods they do not easily scale to higher accuracy regimes. 

% \textbf{more domain decomposition here or how prsent work fits in? whats after this point are remaining notes that you can fill in if still necessary for the overall flow.} 
% Analytic methods are good but targeted. "kernel-independent FMMs" are inducing point methods in disguise, carry the same baggage (although selection of inducing points motivated in some cases by Green's theorem-esque results). RFF will not generate low rank high accuracy results due to the monte carlo nature, better suited for high dimensional problems where low-rank is impossible anyways. RFF can also not do indefinite kernels. 

% Foray into domain decomposition discussion - distinction between on- and off- diagonal compression and why the former doesn't work for some kernels. Discussion on hierarchical methods, matrix sparsity for some kernels and lengthscales (see MEKA paper for example). Point out that our method can be injected into any hierarchical method, for example barnes hut for t-sne. 

% Not sure where to put:
% At some point should mention work on direct solvers. In theory, our routine could be injected into hierarchical routine for direct solve (like skeletonization) if a scaled identity is added and SMW can be used. Also should mention determinant calculation (sylvester determinant theorem). If wanting for content, the exact kernel matrix operations for the kernel methods (GP regression, ridge, etc) can be touched upon. Related work section of fast direct methods for gaussian processes is a smashing good reference for all of the above, although perhaps old info. 
\section{Methods}
\label{sec:methods}
\subsection{Overview}
Here we derive our analytic expansion and the subsequent additional compression step which comprise our new algorithm, which we refer to as a Harmonic Decomposition Factorization (HDF). We will begin by showing why an analytic expansion can lead to a low-rank approximation of a kernel matrix. Then we will show how a Chebyshev expansion of the kernel function can be expanded and rearranged into an expansion in hyperspherical harmonics. From there, our additional compression step examines the functions which form the coefficients in the harmonic expansion\textemdash when evaluated on the data points, these functions can be further reduced in number, thus reducing the overall rank of the factorization. Finally we present error and complexity analyses for the factorization. 
\subsection{Methodology}
We achieve a linear time low-rank approximation by analyzing the kernel's functional form rather than algebraically operating on its associated matrix. Let $K$ be the kernel matrix whose entry in the $i$th row and $j$th column is $k(\|x_i-y_j\|)$ where $x_i, y_j$ are $d$-dimensional points in datasets $\mathcal{X}, \mathcal{Y}$ respectively, and $k$ is a piecewise smooth and continuous isotropic kernel. Our goal is to find $N\times r$ matrices $U, V$ with $r\ll N$ so that the $N\times N$ kernel matrix may be approximated by
\begin{equation}\label{eq:lowrank}K\approx UV^T.\end{equation}
We achieve this by finding functions $u_l, v_l$ so that
\begin{equation}\label{eq:lowrankfunction}
k(\|x-y\|) \approx \sum_{l=0}^{r-1} u_l(x)v_l(y)
\end{equation}
Then we may define $U$ and $V$ as
\[
U_{il}\coloneqq u_l(x_i)\qquad V_{il} \coloneqq v_l(y_i).
\]
This is the typical mechanism of analytic methods for kernel matrix approximation\textemdash $U$ and $V$ are formed by analyzing the kernel function and applying related functions to the data, rather than by examining the original kernel matrix.
% \footnote{This places our method closest to the fast multipole method~\cite{greengard1987fast} and the fast Gauss transform~\cite{greengard1991fast,yang2003improved}. However, unlike those methods we are able to treat a broad class of kernels---in this way our method is closest to~\cite{ryan2021fast}.} 
Thus our main goal is to find $u_l,v_l$ so that~\eqref{eq:lowrankfunction} can be computed efficiently and is sufficiently accurate for as small $r$ as possible. 
\subsection{Analytic Expansion}
First we assume without loss of generality that ${0\leq \|x_i-y_j\|\leq 1}$ for all $x_i,y_j$.\footnote{Mathematically, we're absorbing the diameter of the point set into a lengthscale parameter in the kernel function.} We begin by defining 
\begin{equation}\label{eq:chebtrunc}
\bar{k}(r) \coloneqq 
\begin{cases}
k(r) & r\geq 0
\\ 
k(-r) & r < 0
\end{cases}
\end{equation}
and then forming a truncated Chebyshev expansion of $\bar{k}(r)$ for $-1\leq r \leq 1$
\[
\bar{k}(r) \approx \sum_{i=0}^p a_i T_i(r),
\]
where $T_i$ is the $i$th Chebyshev polynomial of the first kind. The coefficients $a_i$ may be computed efficiently and accurately using \emph{e.g.} the discrete cosine transform, see \cite{clenshaw1960curtis}. This yields
\begin{equation}\label{eq:expbegin}
k(\|x - y\|) \approx \sum_{i=0}^p a_i T_i(\|x-y\|)
\end{equation}
for $0\leq \|x-y\|\leq 1$. Furthermore $a_i=0$ for odd $i$ since $\bar{k}$ is even by definition. To see this, note that
${a_i\coloneqq \int_{-1}^1\bar{k}(r)T_i(r)\mathrm{d}r}$,
and when $i$ is odd the integrand is an even function times an odd function, hence the integral vanishes. Consequently, we henceforth assume that $p$ is even. 

The coefficients in the Chebyshev polynomials are defined as $t_{i,j}$ so that
\begin{equation}\label{eq:chebdef}
T_i(\|x-y\|) = \sum_{j=0}^i t_{i,j}\|x-y\|^{j}
\end{equation}
and, by definition of the Chebyshev polynomials,  
\[
t_{i,j} =
\begin{cases}
1 & i=j \\ 
-t_{i-2, j} & j=0 \\
2t_{i-1, j-1}-t_{i-2,j} & i \neq j, j>0.
\end{cases}
\]
Noting $\|x-y\|^2 = \|x\|^2+\|y\|^2-2x^Ty$ and plugging~\eqref{eq:chebdef} into~\eqref{eq:expbegin} yields
\[
k(\|x-y\|) \approx
\sum_{i=0}^p
\sum_{j=0}^i 
a_i 
t_{i,j} 
(\|x\|^2+\|y\|^2-2x^Ty)^{j/2},
\]
Applying the multinomial theorem and rearranging the sums (see Section~\ref{sec:derivation} for details) results in
\begin{align}
\label{eq:pregegen}
\nonumber
&k(\|x-y\|) \approx\\
&\sum_{k_3=0}^{p/2}
\sum_{k_2=0}^{p/2-k_3}
\sum_{k_1=0}^{p/2-k_3-k_2}
(x^Ty)^{k_3}
\|y\|^{2k_2}
\|x\|^{2k_1}
\mathcal{T}_{k_1, k_2, k_3}
\end{align}
where we have absorbed $a_i, t_{i,j}$, and the multinomial coefficients into constants $\mathcal{T}_{k_1, k_2, k_3}$.

Since $x^Ty = \|x\|\|y\|\cos{\gamma}$ where $\gamma$ is the angle between the vectors $x,y$, we may write
\[
(x^Ty)^{k_3}
\|y\|^{2k_2}
\|x\|^{2k_1}
=
(\cos{\gamma})^{k_3}
\|y\|^{2k_2-k_3}
\|x\|^{2k_1-k_3}.
\]
Plugging this into~\eqref{eq:pregegen} and performing additional rearranging/reindexing yields 
\begin{equation}\label{eq:preharm}
k(\|x-y\|)\approx
\sum_{k=0}^{p/2}
C_k^{\alpha}(\cos{\gamma})
\sum_{m=k}^{p-k}
\sum_{n=k}^{p-m}
\|y\|^{m}
\|x\|^{n}
\mathcal{T}^{'}_{k,m,n}
\end{equation}
where $\alpha =\frac{d}{2}-1$, $C_j^{\alpha}(\cos{\gamma})$ is the Gegenbauer polynomial of the $j$th order, and $\mathcal{T}^{'}_{j,m,n}$ are constants which depend on the kernel but not on $x,y$.

We may finally put this in the form of~\eqref{eq:lowrankfunction} by using the hyperspherical harmonic addition theorem~\citep{averyproperties}:
 \begin{equation}\label{eq:additiontheorem}
 \frac{1}{Z_k^{(\alpha)}}C_k^{(\alpha)}(\cos{\gamma})
 = \sum_{h\in \mathcal{H}_k}
 \Upsilon_k^h(x)
 \Upsilon_k^h(y)
 \end{equation}
 where $Z_k^{(\alpha)}$ is a normalization term, $\Upsilon_k^h$ are hyperspherical harmonics\footnote{We use a real basis of hyperspherical harmonics to avoid complex arithmetic.} which are $d$-dimensional generalizations of the spherical harmonics (see \citep{averyhypharm} for their definition) and  \[
 \mathcal{H}_k\coloneqq \{(\mu_1, \dots \mu_{d-2}) : k\geq\mu_1 \geq \dots \geq |\mu_{d-2}|\geq 0\}.
 \]
 Substituting~\eqref{eq:additiontheorem} into~\eqref{eq:preharm} yields
 \begin{align}\label{eq:preeigen}
 \nonumber
&k(\|x-y\|) \approx\\
&
\sum_{k=0}^{p/2}
\sum_{h\in \mathcal{H}_k}
  \Upsilon_k^h(x)
 \Upsilon_k^h(y)
\sum_{m=k}^{p-k}
 \|y\|^{m}
 \sum_{n=k}^{p-m}
\|x\|^{n}
\mathcal{T}^{''}_{k,m,n}
.\end{align}

This is of the form in~\eqref{eq:lowrankfunction}, using
\[
u_{k,h,m}(x) = \Upsilon_k^h(x) \sum_{n=k}^{p-m}
\|x\|^{n}
\mathcal{T}^{''}_{k,m,n}
\]\[ 
v_{k,h,m}(y) =  \Upsilon_k^h(y)
 \|y\|^{m}.
\]

\begin{algorithm}[t]
\caption{Harmonic Decomposition Factorization (HDF)}
\begin{algorithmic}[t]
\STATE{\textbf{Input}: kernel, error tolerance $\varepsilon$, datasets $x,y$. }
\STATE{\textbf{Output}: matrices $U, V$ s.t. $K\approx UV^T$}
\STATE{$p \gets$ ChebDegreeNeeded($kernel, \varepsilon$)} \quad \COMMENT{Remark~\ref{remark}}
\STATE{$ai \gets$ ChebyshevTransform($kernel,p$)}
\STATE{$idx\gets 0$}
\FOR{$k \in$ 0..$p/2$}
    \STATE{$X^{(k)} \gets$ InitX($k,ai,x$)} \quad\COMMENT{Eq.~\eqref{eq:vandermonde}}
    \STATE{$Y^{(k)} \gets$ InitY($k,y$)}
    \STATE{$Q_X^{(k)}, R_X^{(k)} \gets$ qr($X^{(k)}$)}
    \STATE{$Q_Y^{(k)}, R_Y^{(k)} \gets$ qr($Y^{(k)}$)}
    \STATE{$U^{(k)}, \Sigma^{(k)}, V^{(k)} \gets$ truncsvd($R_X^{(k)}(R_Y^{(k)})^T, \varepsilon$)}
    \STATE{$s_k\gets $numcols($\Sigma^{(k)}$)}
           \IF{$s_k>0$}
\STATE{$\bar{X}^{(k)} \gets Q_X^{(k)}U^{(k)}\sqrt{\Sigma^{(k)}}$}
      \STATE{$\bar{Y}^{(k)} \gets Q_Y^{(k)}V^{(k)}\sqrt{\Sigma^{(k)}}$}
      \FOR{$h\in\mathcal{H}_k$}
      \STATE{$(\Upsilon_k^h)_X \gets$ Harmonic($k,h, x)$}
      \STATE{$(\Upsilon_k^h)_Y \gets$ Harmonic($k,h, y)$}
      \FOR{$l\in 1..s$}
      \STATE{$U[:, idx]\gets \bar{X}^{(k)}\odot (\Upsilon_k^h)_X$}
      \STATE{$V[:, idx]\gets \bar{Y}^{(k)}\odot (\Upsilon_k^h)_Y$}
      \STATE{$idx \gets idx+1$}
      \ENDFOR
\ENDFOR
\ENDIF
\ENDFOR
\STATE{\textbf{return} $U, V$}
\end{algorithmic}\label{alg:fact}
\end{algorithm} 

\subsection{Additional Data-Dependent Compression}
The number of functions in the approximate expansion (which is $r$ in~\eqref{eq:lowrankfunction}) will henceforth be referred to as the \emph{rank} of the expansion. The cost of computing the low-rank expansion is the cost of populating the matrices $U$ and $V^T$, which is linear in the number of entries in the matrices. To achieve efficiency, we'd like the rank of the expansion to be as small as possible while still achieving the desired level of accuracy. Unfortunately, the rank of~\eqref{eq:preeigen} is still large\textemdash proportional to $d^p$. In this section, we incorporate the datasets $\mathcal{X},\mathcal{Y}$ into a scheme that efficiently reduces the rank of~\eqref{eq:preeigen}.

First, we define 
\begin{equation}\label{eq:rdef}
r^{(k)}(\|x\|,\|y\|) \coloneqq \sum_{m=k}^{p-k} \|y\|^m \sum_{n=k}^{p-m} \|x\|^n\mathcal{T}^{''}_{k,m.n} 
\end{equation}
so that~\eqref{eq:preeigen} may be written as
\[k(\|x-y\|)\approx \sum_{k=0}^{p/2}
\sum_{h\in\mathcal{H}_k} \Upsilon_k^h(x) \Upsilon_k^h(y) 
r^{(k)}(\|x\|,\|y\|).
\]
The function $r^{(k)}$ has rank $\frac{p-2k}{2}$,\footnote{$\mathcal{T}^{''}_{k,m,n}$ is nonzero only if $k,m,n$ have the same parity, see Section~\ref{sec:derivation}.} but if we can find an approximation $\tilde{r}^{(k)}\approx r^{(k)}$ with smaller rank and within our error tolerance (when evaluated on the data), then it may be substituted into~\eqref{eq:preeigen} to reduce the overall rank. 

We will find this low-rank approximation by efficiently computing a low-rank approximation of the associated  $N\times N$ matrix defined by
${
(\mathcal{R}^{(k)})_{ij} \coloneqq r^{(k)}(\|x_i\|,\|y_j\|).
}$
Notably, $\mathcal{R}^{(k)}$ can be written as
${
\mathcal{R}^{(k)} =X^{(k)}(Y^{(k)})^T,
}$
where
\begin{equation}\label{eq:vandermonde}
X^{(k)}_{il} = \sum_{n=k}^{p-k-2l}\|x_i\|^n\mathcal{T}^{''}_{k,k+2l, n},\qquad 
Y^{(k)}_{il} = \|y_i\|^{k+2l}.
\end{equation}
Letting $X^{(k)} = Q^{(k)}_XR^{(k)}_X$ and $Y^{(k)} = Q^{(k)}_YR^{(k)}_Y$ be $QR$ factorizations we have
\begin{equation}\label{eq:svd}
\mathcal{R}^{(k)} =Q^{(k)}_X U^{(k)}\Sigma^{(k)} (V^{(k)})^T (Q^{(k)}_Y)^T
\end{equation}
where $U^{(k)}\Sigma^{(k)} (V^{(k)})^T$ is an SVD of the  matrix $R^{(k)}_X(R^{(k)}_Y)^T$. Using an appropriate error tolerance $\tau$, we may find a lower rank representation of $\mathcal{R}^{(k)}$ by truncating all singular values of $\Sigma^{(k)}$ less than $\tau$. This results in 
\begin{equation}\label{eq:rapprox}
{\mathcal{R}^{(k)} \approx \bar{X}^{(k)}(\bar{Y}^{(k)})^T}\end{equation} where
\[
\bar{X}^{(k)}\coloneqq Q^{(k)}_X U^{(k)}_{:,1:s_k}\sqrt{\Sigma^{(k)}_{1:s_k,1:s_k}} \]\[
\bar{Y}^{(k)} \coloneqq Q^{(k)}_Y V^{(k)}_{:,1:s_k}\sqrt{\Sigma^{(k)}_{1:s_k,1:s_k}},
\]
where the notation $U^{(k)}_{:,1:s_k}$ refers to the first $s_k$ columns of $U^{(k)}$, $\Sigma^{(k)}_{1:s_k,1:s_k}$ means the first $s_k$ rows and columns of $\Sigma^{(k)}$, and $s_k$ is the index of the smallest singular value in $\Sigma^{(k)}$ greater than $\tau$.
The selection of $\tau$ is discussed in Section~\ref{sec:error}.
\begin{remark}
When $\mathcal{X}=\mathcal{Y}$, the matrix $\mathcal{R}^{(k)}$ is symmetric, hence the SVD may be replaced with an eigendecomposition. In fact, only the $QR$ factorization of $Y^{(k)}$ is needed in this case. Furthermore, we will have $U=V$ in~\eqref{eq:lowrank}, so the memory requirements are halved. From a practical point of view, it is efficient to include a diagonal matrix $D$ containing these eigenvalues so that we may store $K\approx UDU^T$ with all real entries instead of $K\approx UU^T$ with possibly complex entries in the case of non-positive definite matrices (see \citep{indefkernel} for examples of indefinite learning tasks).
\end{remark}
\begin{remark}
Since the hyperspherical harmonics are orthogonal functions, using the same truncation criteria for different $\mathcal{R}^{(k)}$ yields approximately the same error for each $k$. In practice, this may result in some $\mathcal{R}^{(k)}$ being completely dropped because even their first singular value is below the threshold\textemdash our interpretation is that the $k$th order harmonics in the expansion are unnecessary to achieve the desired level of accuracy, and so this additional compression step allows us to drop them and greatly reduce the rank of the approximation. 
\end{remark}

\subsection{Error}\label{sec:error}
Error is introduced into our approximation in two ways: the truncation of the Chebyshev expansion in~\eqref{eq:chebtrunc} and the data-driven approximation to $\mathcal{R}^{(k)}$ in~\eqref{eq:rapprox}. Suppose we use a threshold of $\tau$ to truncate singuar values/vectors in~\eqref{eq:svd}, then the error $\varepsilon$ of the approximation for any pair of points will satisfy
\begin{multline*}
    \varepsilon =
    \sum_{k=0}^{p/2}
\frac{C_k^{(\alpha)}(\cos{\gamma_{ij}})}{Z_k^{(\alpha)}}
\left(\mathcal{R}^{(k)} - \bar{X}^{(k)}(\bar{Y}^{(k)})^T \right)_{ij}
\\ + \sum_{i=p+1}^\infty a_i T_i(\|x_i-y_j\|).
\end{multline*}
For the first part, note that ${|C_k^{(\alpha)}(\cos{\gamma_{ij}})|\leq \binom{k+d-3}{k}}$ (see \citep{gegenbound} Eq (1.1)). Further, since ${\|A\|_\infty\leq \sqrt{N}\|A\|_2}$ for any $N\times N$ matrix $A$, we have
\[\left|
\left(\mathcal{R}^{(k)} - \bar{X}^{(k)}(\bar{Y}^{(k)})^T \right)_{ij}\right|<\sqrt{N}\tau.\]
For the second part, \cite{elliottpaget} gives
\[\varepsilon_{c}(p)\coloneqq \left|\sum_{i=p+1}^\infty a_i T_i(\|x_i-y_j\|)\right|
\]\[\leq
\frac{1}{2^p(p+1)!}\max_{0\leq r\leq 1}\left|k^{(p+1)}(r)\right|.
\]
Therefore, the total error satisfies
\begin{equation}\label{eq:bound}
|\varepsilon| \leq \sqrt{N} \mathcal{C}_{p/2}\tau + \varepsilon_{c}(p),\end{equation}
where ${
\mathcal{C}_{p/2}
\coloneqq
\sum_{k=0}^{p/2} \binom{k+d-3}{k}/Z_k^{(\alpha)}}.$
\begin{remark}\label{remark}
In practice, $\varepsilon_{c}(p)$ may be easily approximated and we may adaptively choose $p$ and $\tau$ given an input error tolerance $\varepsilon$ such that the desired accuracy is achieved. In fact, since~\eqref{eq:bound} is often quite loose, tuning how $\tau$ is chosen with respect to the input error tolerance can yield significantly smaller ranks while maintaining desired accuracy. 
\end{remark}

\subsection{Complexity} 
The algorithm is summarized in Algorithm~\ref{alg:fact}. Note that $p$ is an adaptively chosen parameter so that~\eqref{eq:chebtrunc} is sufficiently accurate. The Chebyshev transform takes $\mathcal{O}(p)$ time via the discrete cosine transform. If an initial multiplication of Vandermonde matrices is performed outside the $k$ loop, the $QR$ factorizations can be performed in $\mathcal{O}(Np)$ time each (see, for example, \citep{Brubeck2021VandermondeWA}) and so contribute $\mathcal{O}(Np^2)$ to the total cost. Forming $\bar{X}^{(k)},\bar{Y}^{(k)}$ takes $\mathcal{O}(Nps_k)$ each, contributing $\mathcal{O}(Np^2s)$ to the total cost, where $s\coloneqq \max_k s_k$.
The SVDs take $\mathcal{O}(p^3)$ time each and so contribute $\mathcal{O}(p^4)$ to the total cost. 

The total number of harmonics needed by our algorithm will be bounded above by the number of harmonics of order less than or equal to ${p/2}$\textemdash in Section~\ref{sec:numharms} we derive this to be
${\mathbf{H}_{p/2,d}= \binom{\frac{p}{2}+d-1}{d-1}+\binom{\frac{p}{2}+d-2}{d-1}
}.$ Each harmonic costs $\mathcal{O}(d)$ time to evaluate per datapoint, hence the cost of computing the necessary harmonics will be $\mathcal{O}(N\mathbf{H}_{p/2,d}d)$. Once they are computed, the cost to populate the $U$ and $V$ matrices will be $\mathcal{O}(Nr)$ where $r$ is the final rank of the approximation. This rank is given by 
\[r = \sum_{k=0}^{p/2}|\mathcal{H}_k|s_k \leq \sum_{k=0}^{p/2}|\mathcal{H}_k|\left(\frac{p-2k}{2}\right).\]
Thus the total runtime complexity for the factorization is ${\mathcal{O}(N(r+\mathbf{H}_{p/2,d}d+p^2s)+p^4)}$. After the factorization, matrix-vector multiplies may be performed in ${\mathcal{O}(Nr)}$ time.

\subsection{Limitations}
We have described an additional data-dependent compression step based on matrices formed out of the polynomials within the analytic expansion. We currently make no attempt to perform further compression based on the observed rank of the harmonic components. This is logical when the number of points is high relative to the dimension\textemdash matrices formed from the harmonics will have little singular value decay owing to the orthogonality of the harmonic functions. However, if the number of dimensions is large relative to the the number of points, then the points are not space filling and the lack of additional compression based on the harmonics will result in poor efficiency, especially as the number of harmonics grows polynomially with the dimension (see Section~\ref{sec:harmsec} for further discussion). 

Vandermonde matrices are notoriously ill-conditioned~\citep{vanderill}, and the accuracy of the $QR$ factorizations in the current presentation of our additional compression step can suffer when $p$ is large unless this is addressed. In our testing, this degradation only arises for very small length-scale parameters and/or very small error tolerances (notably, outside the range seen in our experiments). One step towards preventing potential conditioning issues could be to arrange the expansion so that Legendre polynomials of $\|x\|,\|y\|$ are used instead of powers of $\|x\|,\|y\|$ (in an analogous way to our replacement of powers of cosine with Gegenbauer polynomials of cosine, see Section~\ref{sec:derivation} for details). 

In many kernel matrix approximation algorithms, a non-negligible cost of generating the low-rank matrices is that matrix entries typically cost $\mathcal{O}(d)$ time to generate. For example, this is the case when a kernel evaluation is required between a point and a benchmark point, as in inducing point methods. On the one hand, our factorization benefits from the $\mathcal{O}(d)$ costs being confined\footnote{The Vandermonde-like matrices require only a single dimension-dependent computation of the norms of all datapoints.} to the computation of the harmonics, each of which is reused across $s_k$ rows where $s_k$ is the rank in~\eqref{eq:rapprox}. On the other hand, this depends on an efficient implementation of hyperspherical harmonic computation\textemdash although we provide a moderately tuned implementation, this is the subject of further work as there are few available good pre-existing routines for this procedure. 

\section{Experiments}\label{sec:experiments}
We have implemented the Harmonic Decomposition factorization in Algorithm~\ref{alg:fact} in Julia as part of an
open source toolkit, and have performed synthetic and real-world regression experiments single-threaded on a 2020 Apple Macbook Air with an M1 CPU and 8GB of RAM. All recorded times are averaged across 10 trials.

\begin{figure}[ht!]
\centering
\includegraphics[width=.8\columnwidth]{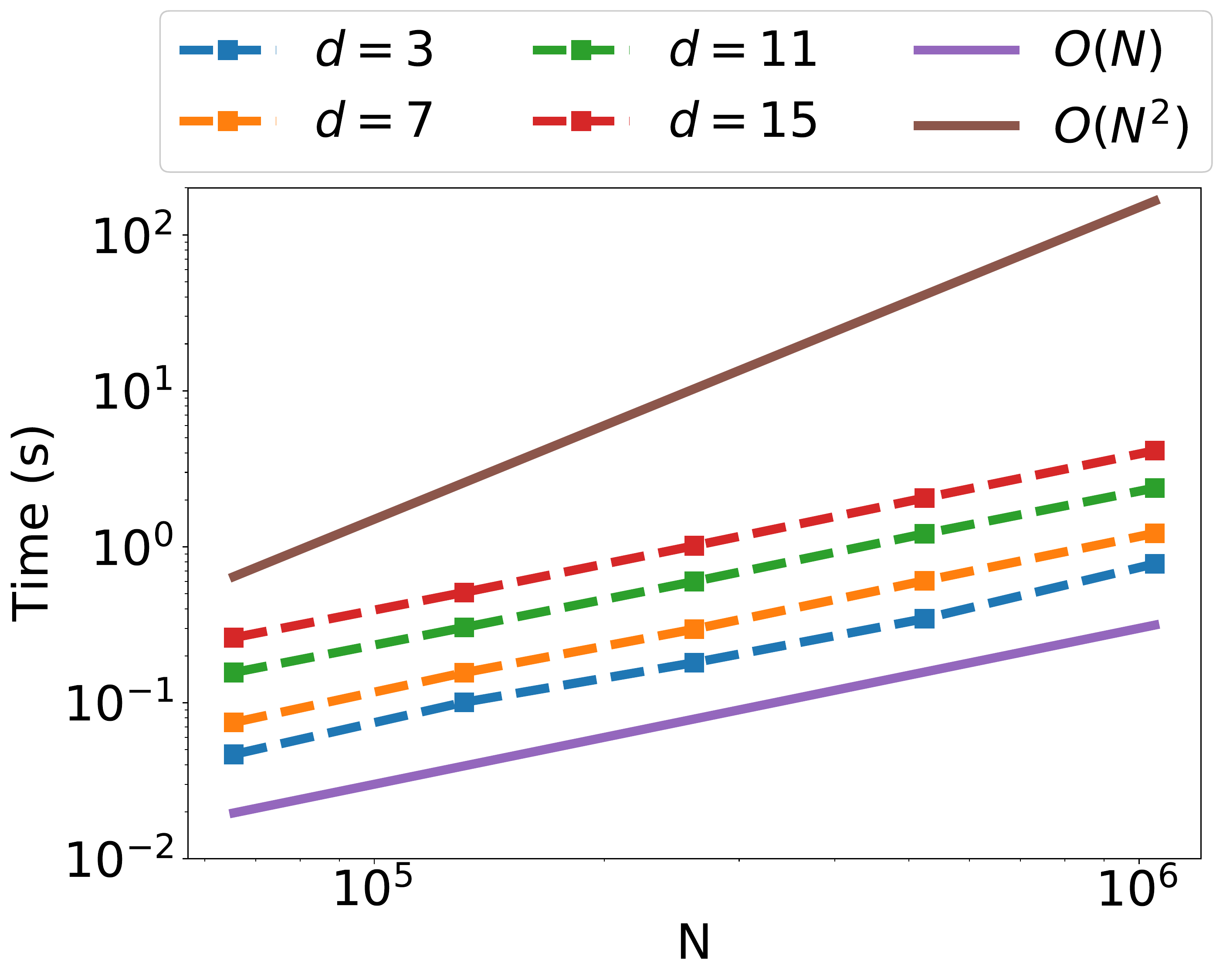}
% }
\vskip -0.2in
\caption{Time vs. dataset size results for the Cauchy kernel with $\sigma=1$. For this experiment we use an error tolerance of $10^{-3}$ and use normally distributed points scaled to have norm less than 1 as our dataset. For a variety of dimensions our algorithm displays linear scaling with the size of the dataset.}
\vskip -0.1in
\label{fig:timevsn}
\end{figure}
\subsection{Synthetic Data}
To test the runtime of our algorithm for constructing the factorization, we generate datasets of points drawn from a normal distribution and subsequently scaled so that the maximum norm of a point in a dataset is $1$. Using the Cauchy kernel with lengthscale parameter $\sigma=1$, we vary the number of points and dimensions of this dataset and record the time required by the implementation to create the low-rank approximation. The relative error tolerance is fixed at $\varepsilon=0.005$. Figure~\ref{fig:timevsn} vizualizes the results of this experiment\textemdash across a variety of dimensions, our factorization requires linear time in the number of points.

\begin{figure}[h!]
\centering
\includegraphics[width=.8\columnwidth]{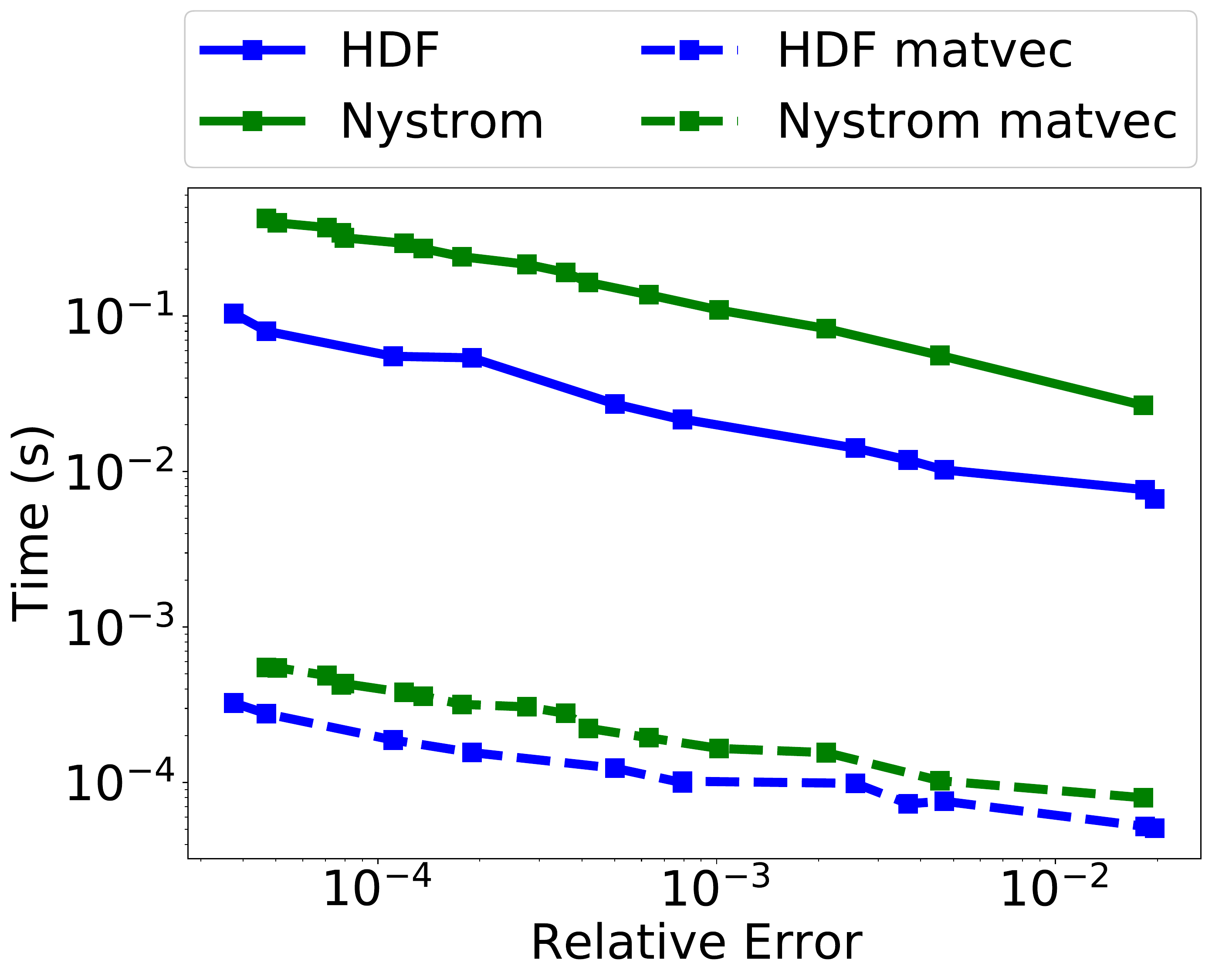}
% }
\caption{Time vs. relative error results for the Cauchy kernel with $\sigma=1$ using a 3D dataset of $10^5$ points generated the same way as in Figure~\ref{fig:timevsn}. For both factorization and matrix-vector multiplies, our method displays a better tradeoff than the Nystr\"{o}m method. }
\label{fig:errvstime}
\end{figure}

To illustrate how efficiently we can tradeoff time and accuracy, we compare our factorization with the popular Nystr\"{o}m method where inducing points are chosen uniformly at random.\footnote{Times for the Nystr\"{o}m method include populating the interpolation matrix as well as factoring the inducing point matrix.} We use a dataset of 10,000 points in three dimensions generated the same as before and using the same Cauchy kernel. This time, the error tolerance is varied and the factorization time and relative errors are recorded. Figure~\ref{fig:errvstime} shows that our factorization is consistently cheaper to produce than the Nystr\"{o}m factorization that achieves the same accuracy. This further translates to faster matrix vector products at the same accuracy. This is due largely to the HDF's lower-rank approximation, which we investigate next. 

\begin{figure*}[ht!]
\centering
\includegraphics[width=.65\columnwidth]{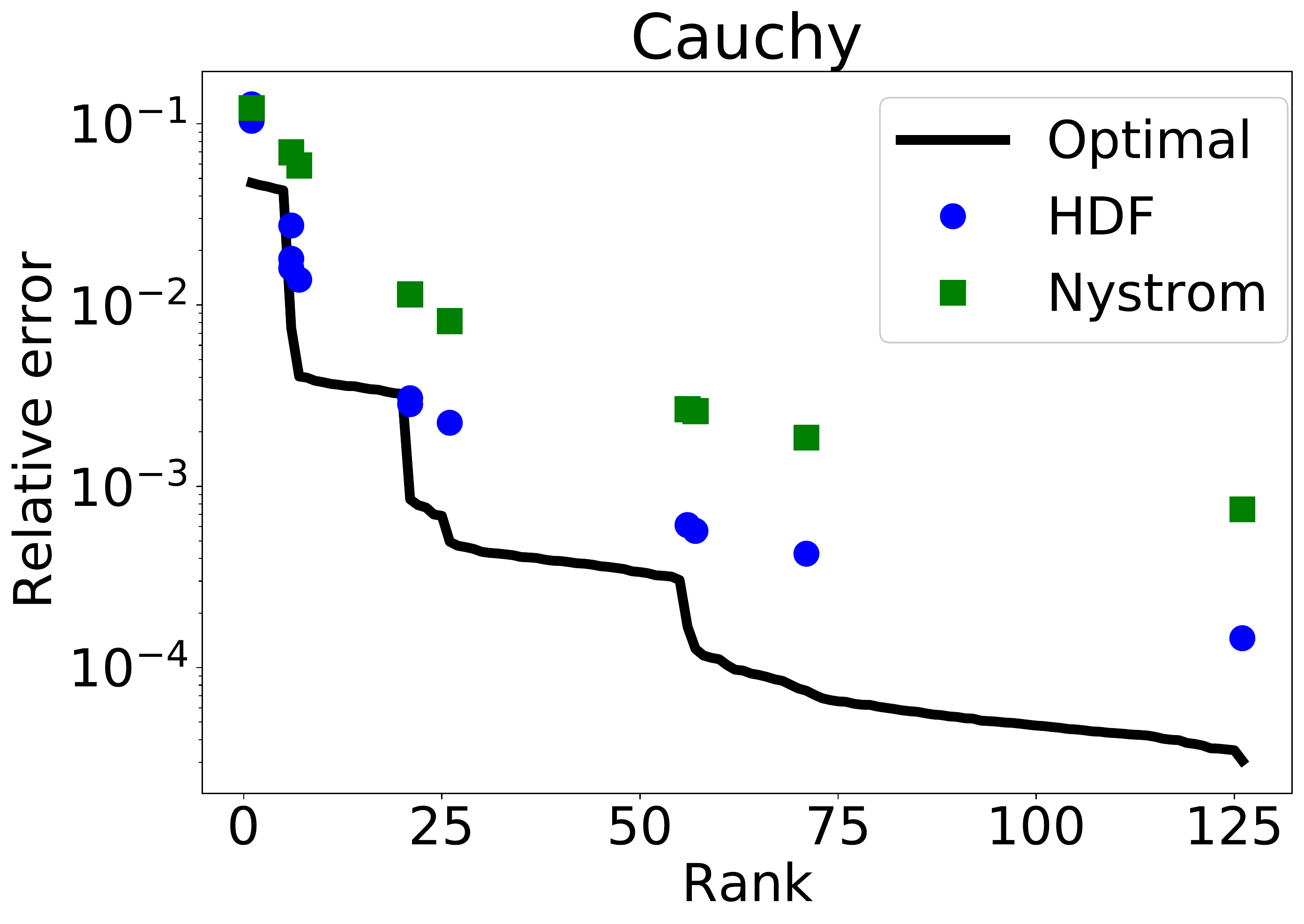}\hspace{15mm}
\includegraphics[width=.65\columnwidth]{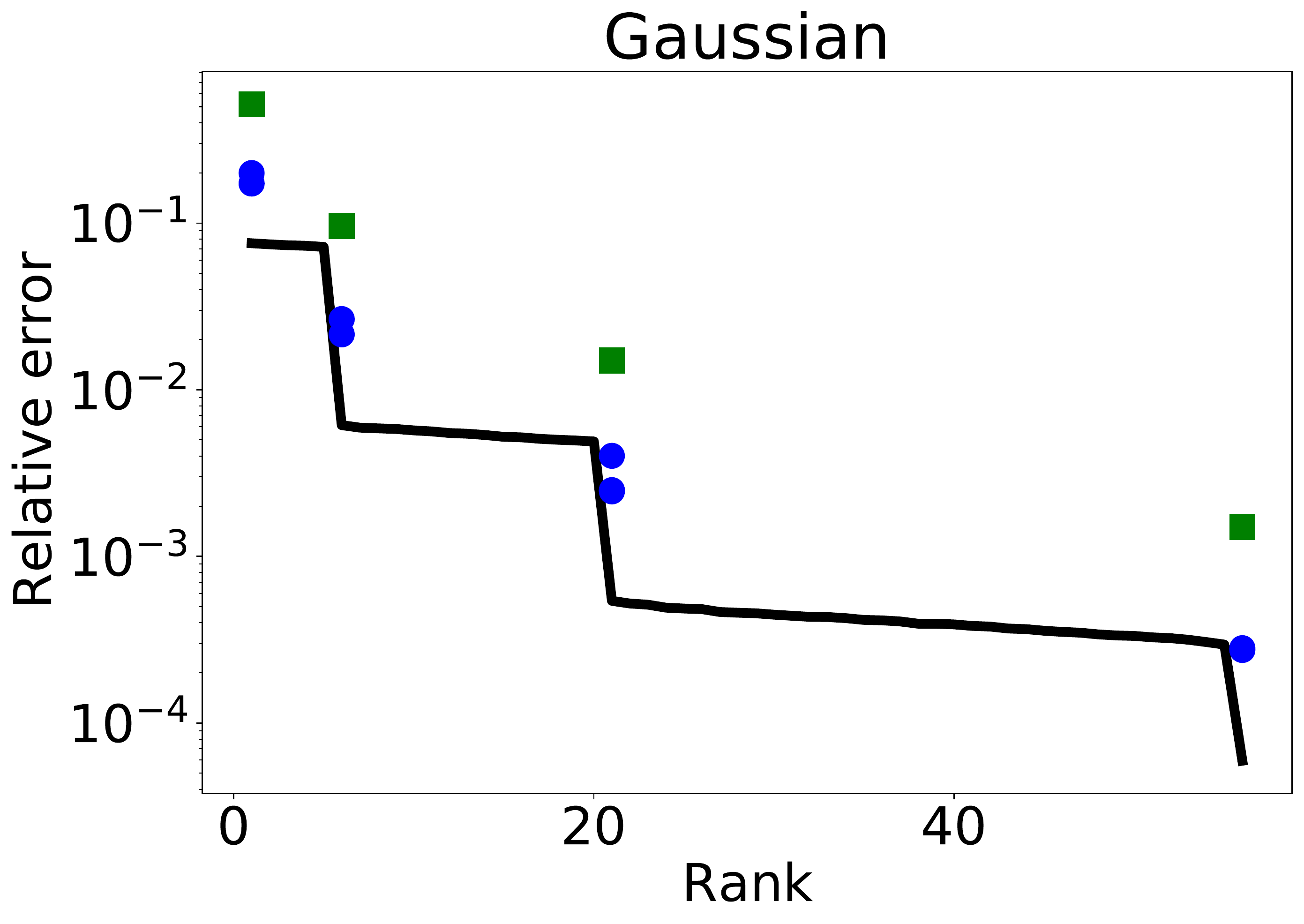}\\
\includegraphics[width=.65\columnwidth]{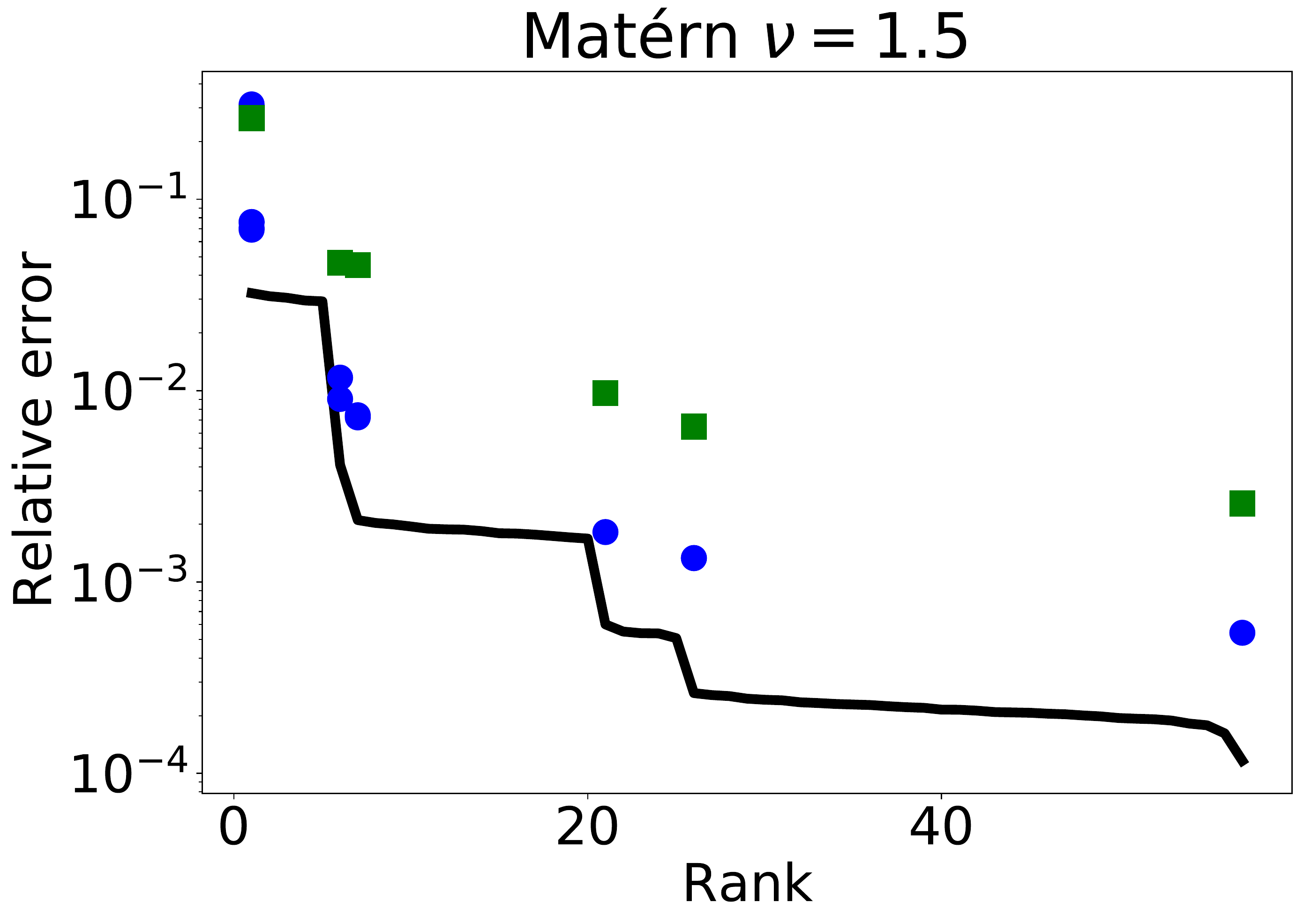}\hspace{15mm}
\includegraphics[width=.65\columnwidth]{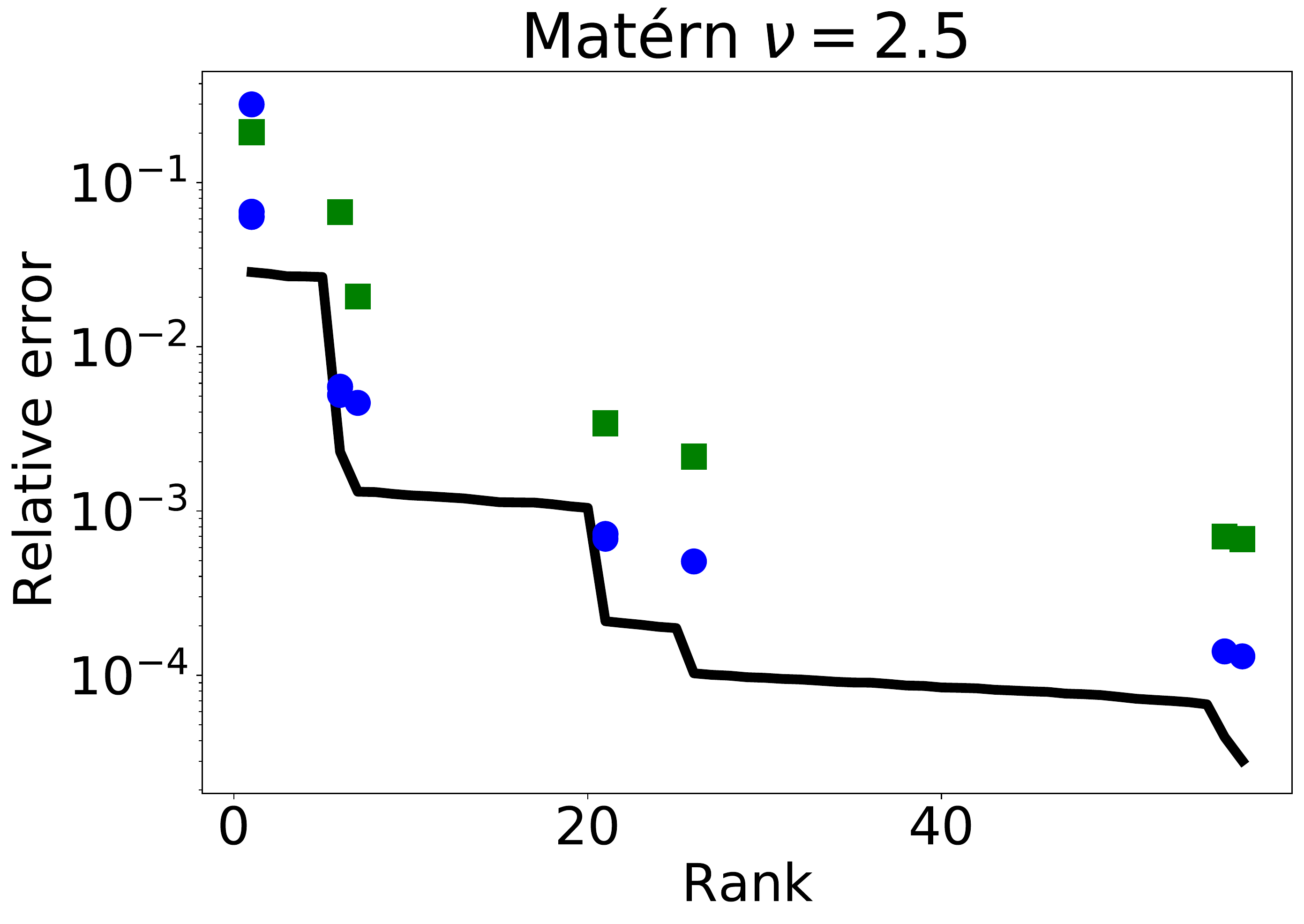}
% }
\vskip -0.2in
\caption{Relative error vs. rank results for a variety of kernels using a 5D dataset of $5000$ points generated the same way as in Figure~\ref{fig:timevsn}. Our method has a substantially improved tradeoff over the Nystr\"{o}m method, and approaches optimality for lower-rank approximations.}\label{fig:relerrvsrank}
\vskip -0.15in
\end{figure*}

To further explore the tradeoff between rank and relative error of our approximation, we conduct experiments where the error tolerance is varied and the rank of the necessary factorization along with the actual final relative error is collected. In this experiment, we use a dataset of 5000 points in five dimensions generated in the same way as above and use the Cauchy kernel $\left((k(r)=\frac{1}{1+(r/\sigma)^2}\right)$, 
Gaussian kernel ($k(r) = e^{-(r/\sigma)^2}$), 
Matérn kernel with $\nu=1.5$ (${k(r)=(1+\sqrt{3}(r/\sigma))e^{-\sqrt{3}r/\sigma}}$)
and Matérn kernel with $\nu=2.5$ (${k(r)=(1+\sqrt{5}(r/\sigma)+(5/3)(r/\sigma)^2)e^{-\sqrt{5}r/\sigma}}$).
Figure~\ref{fig:relerrvsrank} compares results between our method, the Nystr\"{o}m method, and the SVD (which provides the optimal rank/error tradeoff). In all cases our method has greater accuracy than the Nystr\"{o}m method for the same rank. Additionally, our method shows nearly optimal performance for lower rank approximations. This experiment also highlights the generality of our method---most existing analytic methods cannot be applied to this breadth of kernels and are, therefore, omitted from the experiment.

\subsection{Kernel Ridge Regression on California Housing Data}
We also apply our factorization within the training process of kernel ridge regression and evaluate the predictions. Kernel ridge regression~\citep{shawe2004kernel} is a kernel method which involves finding a weight vector $\mathbf{w}$ that solves the minimization problem
\[
\mathbf{w} \coloneqq \argmin_\mathbf{w} \|\mathbf{y}-K(\mathbf{x},\mathbf{x})\mathbf{w}\|^2 + \lambda \mathbf{w}^TK(\mathbf{x},\mathbf{x})\mathbf{w},
\]
where $\mathbf{x}$ are the training points, $\mathbf{y}$ are the training labels, and $\lambda$ is a regularization parameter. The solution to this problem is 
\begin{equation}\label{eq:opt}
    \mathbf{w}=(\lambda\mathbf{I}+K(\mathbf{x},\mathbf{x}))^{-1}\mathbf{y}
\end{equation}
Predictions on test data are then performed using
${\hat{\mathbf{y}} = K(\mathbf{x}^*, \mathbf{x})\mathbf{w}}$,
where $\mathbf{x}^*$ are the test points and $\hat{\mathbf{y}}$ are our predictions of the test labels. 

We use the California housing data set from the UCI Machine Learning Repository \citep{uci}, which contains data drawn from the 1990 U.S. Census. The labels are the median house values, and the features are all other attributes, leading to a feature space of $d=8$ dimensions for a dataset of ${N=20,640}$ points. We scale features to be in $[0,1]$, and for each of five trials perform a random shuffle of the data followed by a $\frac{2}{3}$-$\frac{1}{3}$ train-test split. 

We apply our factorization to the problem by generating low-rank approximations for the diagonal blocks of the kernel matrix in~\eqref{eq:opt}. The diagonal blocks are chosen to be associated with 30 clusters $C_i$ of points generated by $k$-means clustering \citep{kmeans}, so that the $i$th block along the diagonal is given by ${B_i\coloneqq K_{C_i, C_i}}$. A host of techniques exist for compressing the off-diagonal blocks \citep{ying2004kernel,ying2006kernel,ambikasaran2015fast,ryan2021fast}, but we perform no off-diagonal compression for this experiment to simplify the implementation. The associated linear system is then solved using the method of conjugate gradients \citep{iterbook,cgbrief}, where the matrix-vector multiplies are accelerated by our compression. As a preconditioner we use a block diagonal matrix whose blocks are the inverses of the blocks of the true kernel matrix associated with the clusters.

\begin{figure}[ht!]
\centering
\includegraphics[width=.8\columnwidth]{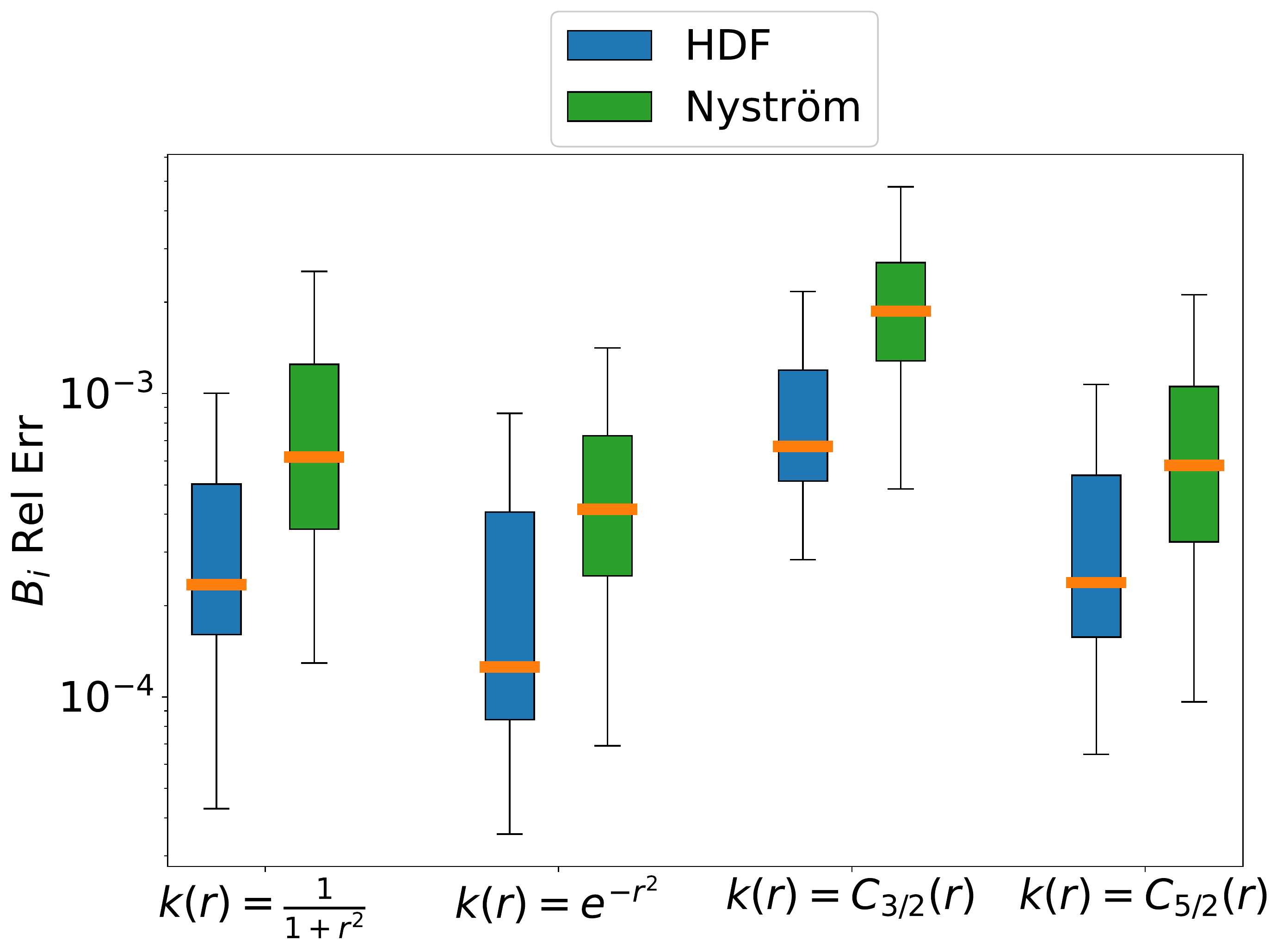}
% }
\vskip -0.15in
\caption{Relative error results for the diagonal block approximations by our method and the Nystr\"{o}m method. The blocks correspond to kernel interactions between points within clusters generated by $k$-means on the California Housing dataset.}
\vskip -0.15in
\label{fig:gpregression}
\end{figure}

For comparison, we also applied the Nystr\"{o}m method using the same ranks for the diagonal blocks. We record the relative errors of the diagonal block approximations, as well as the final mean squared errors (MSEs) when compared with the ground truth test labels. Figure~\ref{fig:gpregression} visualizes our findings. For all kernels, our method tends to find a more accurate approximation than Nystr\"{o}m for the same rank, as also demonstrated for the synthetic data in Figure~\ref{fig:relerrvsrank}.  Further, owing to the more accurate approximations, our method yields MSEs closer to those found using dense operations. 

\section{Conclusion}
We have presented a new method for low-rank approximation of kernel matrices that pairs a novel analytic expansion with an efficient data-dependent compression step. The technique works for any isotropic kernel, and is particularly efficient when the dimension of the data is small relative to the size of the dataset. Our method runs in linear time and has a configurable error tolerance along with theoretical guarantees on the size of the error. This work presents results from a suite of experiments demonstrating linear scaling and favorable comparative metrics against the Nystr\"{o}m method, both for synthetic and real-world data. An open-source implementation of our method is included alongside this paper. 
This work also leads to many interesting and promising questions for future exploration. One currently unexplored facet is whether an additional compression step may be applied to the harmonics. For our areas of application this has not been necessary, as the assumption of moderate dimension has meant that the matrices of harmonics are not numerically low rank. However, for high $d$ and high desired accuracy, the high orders of harmonics required are too numerous for an efficient low-rank approximation (see Figure~\ref{fig:harmrank} in Section~\ref{sec:harmsec}). This problem could potentially be mitigated by a similar data-dependent compression technique as is used in this work, or another dimension-reduction technique. 
\begin{table}[h!]
\begin{center}
    \begin{tabular}{ |l|l|l| }%
      \hline%
       & $\frac{1}{1+(r/\sigma)^2}$
      &$e^{-(r/\sigma)^2}$
       \\
      \hline 
      Dense &  1.72 [1.62, 1.77]e-2 &   1.76 [1.71, 1.80]e-2 \\ %
      HDF &  1.71 [1.62, 1.77]e-2  &   1.76 [1.70, 1.80]e-2   \\ %
    Nyst. &  1.78 [1.71, 2.08]e-2  &   1.79 [1.71, 1.83]e-2 \\ %
      \hline%
      \hline%
       & Matérn $\nu=1.5$
      &  Matérn $\nu=2.5$
       \\
      \hline 
            Dense &  1.58 [1.54,1.62]e-2  &  1.69 [1.66,1.73]e-2  \\ %
      HDF &  1.58 [1.55,1.62]e-2  &  1.69 [1.65,1.73]e-2  \\ %
    Nyst. &  2.52 [1.70,7.94]e-2  &  1.69 [1.68,1.72]e-2  \\ %
    \hline
    \end{tabular}
\end{center}
\caption{Mean squared error (MSE) of predictions compared to ground truth values in the test split of our regression experiment. Entries are the median, minimum, and maximum MSEs observed. In all cases, the higher accuracy of our approximation resulted in better MSE. For the Matérn kernel with $\nu=1.5$ the Nystr\"{o}m approximation yielded conditioning issues, and the iterative method sometimes failed to converge in 1024 iterations.}
\vskip -0.1in
\label{table:mse}
\end{table}

Our expansion was formed by transforming to a basis of Gegenbauer polynomials in $\cos{\gamma}$ and then splitting into harmonics. This approach is inspired by the analytic expansions underlying fast multipole methods, but may not be ideal for high dimensional problems if the hyperspherical harmonics are too onerous to compute. For higher dimensional problems it is conceivable that a more manageable basis for the angular functions would be preferred for a fast implementation. A different such basis could also be engineered to allow for efficiency in an additional compression as described in the previous paragraph, to parallel how the Vandermonde structure of~\eqref{eq:vandermonde} allows for efficiency in our additional compression step. 

We have provided the theoretical underpinnings for off-diagonal compression based on our scheme and future work will incorporate our routine into a hierarchical domain decomposition framework. It is plausible that an opportunity exists to reuse matrices or at least harmonic computations between on- and off-diagonal block approximations.

\bibliography{hdf}
\bibliographystyle{icml2022}
\onecolumn
\appendix
\section{Derivation of the Analytic Expansion}\label{sec:derivation}
Here we present a detailed derivation of the analytic expansion which forms the foundation of our algorithm. Starting from the Chebyshev expansion
\[
k(\|x - y\|) \approx \sum_{i=0}^p a_i T_i(\|x-y\|)
\]
Using the definition of the Chebyshev polynomial \[T_i(\|x-y\|) = \sum_{j=0}^i t_{i,j}\|x-y\|^{j},\]
we get
\[k(\|x - y\|) \approx 
\sum_{i=0}^p
\sum_{j=0}^i 
a_i 
t_{i,j} 
(\|x\|^2+\|y\|^2-2x^Ty)^{j/2}.
\]
Applying the multinomial theorem yields
\[k(\|x - y\|) \approx 
\sum_{i=0}^p
\sum_{j=0}^i 
\sum_{k_3=0}^{j/2}
\sum_{k_2=0}^{j-k_3}
t_{i,j}
\|x\|^{j-2k_3-2k_2}
\|y\|^{2k_2}
(-2x^Ty)^{k_3}
\binom{j/2}{j/2-k_3-k_2, k_2, k_3}
a_i.
\]

Swapping sums via
\[
\sum_{i=0}^p
\sum_{j=0}^i 
\sum_{k_3=0}^{j/2}
\sum_{k_2=0}^{j/2-k_3}
=
\sum_{i=0}^p
\sum_{k_3=0}^{i/2}
\sum_{j=2k_3}^i 
\sum_{k_2=0}^{j/2-k_3}
=
\sum_{k_3=0}^{p/2}
\sum_{i=2k_3}^p
\sum_{j=2k_3}^i 
\sum_{k_2=0}^{j/2-k_3}
\]
\[=
\sum_{k_3=0}^{p/2}
\sum_{i=2k_3}^p
\sum_{k_2=0}^{i/2-k_3}
\sum_{j=2k_2+2k_3}^i 
=
\sum_{k_3=0}^{p/2}
\sum_{k_2=0}^{p/2-k_3}
\sum_{i=2k_2+2k_3}^p
\sum_{j=2k_2+2k_3}^i 
\]
gives

\[k(\|x - y\|) \approx 
\sum_{k_3=0}^{p/2}
(x^Ty)^{k_3}
\sum_{k_2=0}^{p/2-k_3}
\|y\|^{2k_2}
\sum_{i=2k_2+2k_3}^p
\sum_{j=2k_2+2k_3}^i 
t_{i,j}
\|x\|^{j-2k_3-2k_2}
(-2)^{k_3}
\binom{j/2}{j/2-k_3-k_2, k_2, k_3}
a_i
\]
Let $k_1\coloneqq j/2-k_3-k_2$ so that $j=2k_1+2k_2+2k_3$, then we have
\[k(\|x - y\|) \approx 
\sum_{k_3=0}^{p/2}
(x^Ty)^{k_3}
\sum_{k_2=0}^{p/2-k_3}
\|y\|^{2k_2}
\sum_{i=2k_2+2k_3}^p
\sum_{k_1=0}^{i/2-k_3-k_2}
t_{i,2k_1+2k_2+2k_3}
\|x\|^{2k_1}
(-2)^{k_3}
\binom{k_1+k_2+k_3}{k_1, k_2, k_3}
a_i
\]
Swapping the $i$ and $k_1$ sums gives
\[k(\|x - y\|) \approx 
\sum_{k_3=0}^{p/2}
(x^Ty)^{k_3}
\sum_{k_2=0}^{p/2-k_3}
\|y\|^{2k_2}
\sum_{k_1=0}^{p/2-k_3-k_2}
\|x\|^{2k_1}
\sum_{i=2k_1+2k_2+2k_3}^p
t_{i,2k_1+2k_2+2k_3}
(-2)^{k_3}
\binom{k_1+k_2+k_3}{k_1, k_2, k_3}
a_i
\]
Using $a\cdot b = \|a\|\|b\|\cos{\gamma}$ gives 
\[k(\|x - y\|) \approx 
\sum_{k_3=0}^{p/2}
(\cos{\gamma})^{k_3}
\sum_{k_2=0}^{p/2-k_3}
\|y\|^{2k_2+k_3}
\sum_{k_1=0}^{p/2-k_3-k_2}
\|x\|^{2k_1+k_3}
\sum_{i=2k_1+2k_2+2k_3}^p
\dots
\]
We will use the identity 
\[\cos^i{\gamma} = \sum_{k=0}^iA_{ki}C_k^{\alpha}(\cos{\gamma})\]
where $\alpha=d/2-1$, $C_k^{\alpha}(\cos{\gamma})$ is the Gegenbauer polynomial, $A_{ki}=0$ when $k\neq i \mod 2$, and
\[
A_{ki} = \frac{i!(\alpha+k)}{2^i
\frac{i-k}{2}
(\alpha)_{ \frac{i+k}{2} + 1 }}
\]
when $k= i \mod 2$. The notation $(\alpha)_{ \frac{i+k}{2} + 1 }$ refers to the rising factorial, i.e. $(\alpha)_n = (\alpha)(\alpha+1)\dots(\alpha+n-1)$. Using this to expand the $(\cos{\gamma})^{k_3}$ term yields
\[k(\|x - y\|) \approx 
\sum_{k_3=0}^{p/2}
\sum_{j=0}^{k_3}
A_{k_3j}
C_j(\cos{\gamma})
\sum_{k_2=0}^{p/2-k_3}
\|y\|^{2k_2+k_3}
\sum_{k_1=0}^{p/2-k_3-k_2}
\|x\|^{2k_1+k_3}
\sum_{i=2k_1+2k_2+2k_3}^p
\dots
\]
Swapping the $k_3$ and $j$ sums yields
\[k(\|x - y\|) \approx 
\sum_{j=0}^{p/2}
C_j(\cos{\gamma})
\sum_{k_3=j}^{p/2}
\sum_{k_2=0}^{p/2-k_3}
A_{k_3j}
\|y\|^{2k_2+k_3}
\sum_{k_1=0}^{p/2-k_3-k_2}
\|x\|^{2k_1+k_3}
\sum_{i=2k_1+2k_2+2k_3}^p
\dots
\]
Let $m=2k_2+k_3$ so that $k_2 = (m-k_3)/2$
% p-k_3
\[k(\|x - y\|) \approx 
\sum_{j=0}^{p/2}
C_j(\cos{\gamma})
\sum_{k_3=j}^{p/2}
\sum_{m=k_3}^{p-k_3}
A_{k_3j}
\|y\|^{m}
\sum_{k_1=0}^{p/2-k_3/2-m/2}
\|x\|^{2k_1+k_3}
\sum_{i=2k_1+m+k_3}^p
\dots
\]
Swapping the $k_3$ and $m$ sums gives
% k3 >= j
% k3 <= p/2
% k3 <= m
% p-k3 >= m 
% meaning k3 <= p-m
\[k(\|x - y\|) \approx 
\sum_{j=0}^{p/2}
C_j(\cos{\gamma})
\sum_{m=j}^{p-j}
\|y\|^{m}
\sum_{k_3=j}^{\min(p/2, m, p-m)}
\sum_{k_1=0}^{p/2-k_3/2-m/2}
A_{k_3j}
\|x\|^{2k_1+k_3}
\sum_{i=2k_1+m+k_3}^p
\dots
\]

Let $n=2k_1+k_3$ so that $k_1 = (n-k_3)/2$

\[k(\|x - y\|) \approx 
\sum_{j=0}^{p/2}
C_j(\cos{\gamma})
\sum_{m=j}^{p-j}
\|y\|^{m}
\sum_{k_3=j}^{\min(p/2, m, p-m)}
\sum_{n=k_3}^{p-m}
A_{k_3j}
\|x\|^{2k_1+k_3}
\sum_{i=2k_1+m+k_3}^p
\dots
\]
Swapping the $k_3$ and $n$ sums yields 
\[k(\|x - y\|) \approx 
\sum_{j=0}^{p/2}
C_j(\cos{\gamma})
\sum_{m=j}^{p-j}
\|y\|^{m}
\sum_{n=j}^{p-m}
\|x\|^{n}
\sum_{k_3=j}^{\min(p/2, m, p-m, n)}
A_{k_3j}
\sum_{i=n+m}^p
\dots
\]
\[
k(\|x - y\|) \approx 
\sum_{j=0}^{p/2}
C_j(\cos{\gamma})
\sum_{m=j}^{p-j}
\|y\|^{m}
\sum_{n=j}^{p-m}
\|x\|^{n}
\mathcal{T}^{'}_{j,m,n}
\]
where 
\[
\mathcal{T}^{'}_{j,m,n}
 = 
\sum_{k_3=j}^{\min(p/2, m, p-m, n)}
A_{k_3j}
\sum_{i=
n+m}^p
t_{i,n+m}
(-2)^{k_3}
\binom{(n+m)/2}{(n-k_3)/2, (m-k_3)/2, k_3}
a_i\]
The final form of the expansion follows from applying the hyperspherical harmonic addition theorem. 

\section{Number of Harmonics of Order $\leq k$}\label{sec:numharms}
Here we show that
\[\mathbf{H}_k\coloneqq\sum_{k=0}^{p/2}|\mathcal{H}_k| =\binom{\frac{p}{2}+d-1}{d-1}+\binom{\frac{p}{2}+d-2}{d-1}.
\]
Note that \citep{averyproperties} gives
\[
|\mathcal{H}_k| = \binom{k+d-1}{k}-\binom{k+d-3}{k-2}
\]
Thus the LHS forms a telescoping sum and we are left with
\[
\binom{\frac{p}{2}+d-1}{\frac{p}{2}}+\binom{\frac{p}{2}+d-2}{\frac{p}{2}-1}  = \binom{\frac{p}{2}+d-1}{d-1}+\binom{\frac{p}{2}+d-2}{d-1} 
\]
\section{Compression of Matrices of Harmonics}\label{sec:harmsec}
\begin{figure}[t]
\centering
\includegraphics[width=0.45\columnwidth]{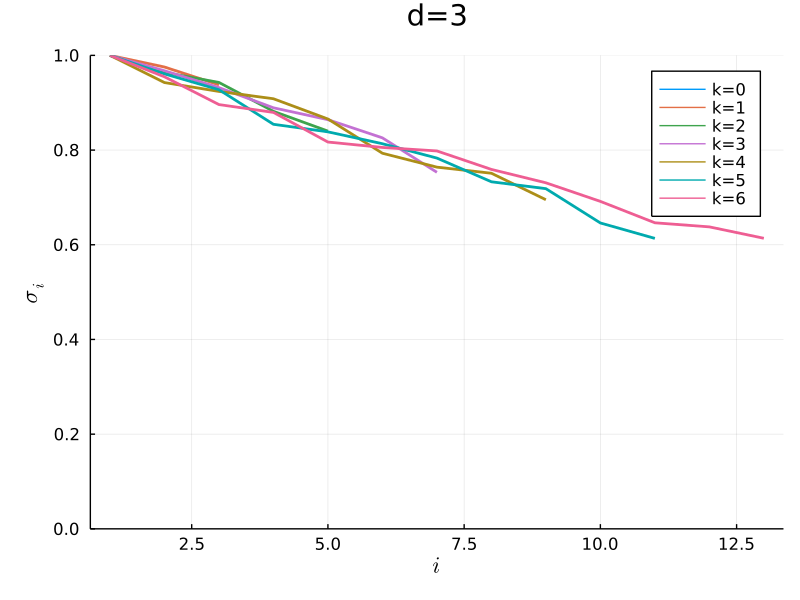}
\includegraphics[width=0.45\columnwidth]{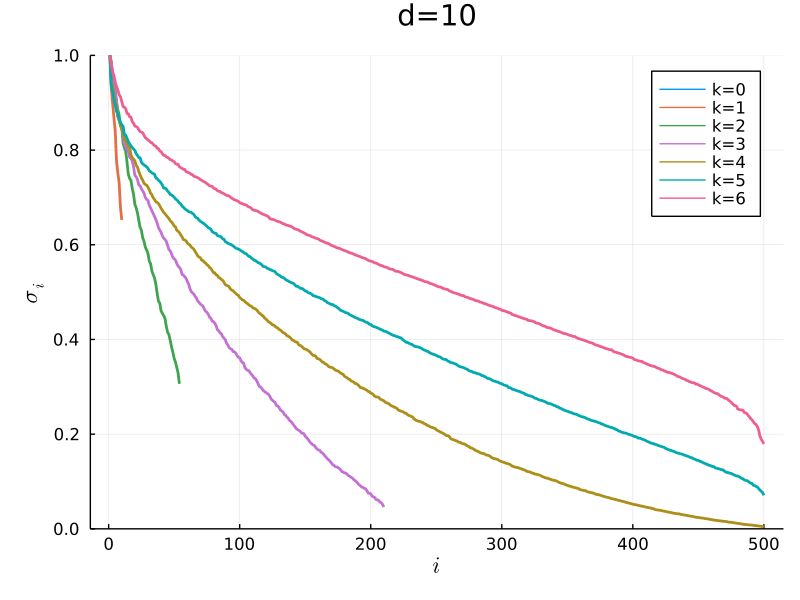}
\caption{The decay of the singular values of the matrices formed out of the harmonics in the expansion. Specifically, the matrices are defined by $M^{(k)}_{ij}\coloneqq \sum_{h\in\mathcal{H}_k} \Upsilon_k^h(x_i)\Upsilon_k^h(x_j)$ for $1\leq i,j\leq N$. This experiment used $N=500$ normally distributed points. For small $d$, there is no real need to compress the harmonic components as the orthogonality of the functions leads to matrices with little singular value decay. However, for high $d$ the number of harmonics grows substantially, and the decay of the singular values of the associate matrices suggests that an additional compression would need to be performed on the harmonic components to remain efficient.}
\label{fig:harmrank}
\end{figure}

The algorithm we presented includes an additional data-dependent compression step which involves finding a low-rank approximation to the matrices $\mathcal{R}^{(k)}$ as in~\eqref{eq:rapprox}. We could, in an analogous fashion, attempt to find a data-dependent compression of the matrix of harmonics defined by
\[
M^{(k)}_{ij}\coloneqq \sum_{h\in\mathcal{H}_k} \Upsilon_k^h(x_i)\Upsilon_k^h(x_j).
\]
For problems where the dimension is large relative to the size of the dataset, this matrix can have fast-decaying singular values, as seen in Figure~\ref{fig:harmrank}, and hence it would be useful to perform additional compression in those cases. Future research will search for efficient ways to perform this compression when it is called for, or prevent the need for it with some suitable data transformation. 

\end{document}